\documentclass[10pt,twocolumn,letterpaper]{article}
\usepackage[table]{xcolor}

\usepackage{amsmath}
\usepackage{eso-pic,xspace}
\usepackage{amssymb}

\usepackage{graphicx}
\usepackage{comment}
\usepackage{amsmath,amssymb} 
\usepackage{color} 

\usepackage{cvpr}
\usepackage{times}
\usepackage{epsfig}
\usepackage{graphicx}
\usepackage{amsmath}
\usepackage{amssymb}
\usepackage{xspace} 
\usepackage{color}
\usepackage{verbatim}
\usepackage{color}
\usepackage{tabularx}
\usepackage{amssymb}
\usepackage{amsmath}
\usepackage{amsmath,amssymb} 
\usepackage{adjustbox}
\usepackage{multirow}
\usepackage{booktabs}
\usepackage[normalem]{ulem}
\usepackage{xfrac}
\usepackage{url} 
\usepackage{xspace} 
\usepackage{wrapfig}

\makeatletter
\@namedef{ver@everyshi.sty}{}
\makeatother
\usepackage{tikz}
\usetikzlibrary{svg.path}
 
\usepackage[breaklinks=true,bookmarks=false]{hyperref}

\cvprfinalcopy 

\def\cvprPaperID{****}

\definecolor{boxGreen}{rgb}{0.44,0.68,0.28}
\definecolor{boxRed}{rgb}{0.75,0.,0.}
\definecolor{boxYellow}{rgb}{1.,0.75,0.}
\definecolor{boxBlue}{rgb}{0.27,0.45,0.77}
\definecolor{boxOrange}{rgb}{0.93,0.49,0.19} 
\definecolor{apricot}{rgb}{0.98, 0.81, 0.69}
\definecolor{asparagus}{rgb}{0.53, 0.66, 0.42}
\definecolor{palerobineggblue}{rgb}{0.59, 0.87, 0.82}
\definecolor{olivine}{rgb}{0.6, 0.73, 0.45}
\definecolor{oldlace}{rgb}{0.99, 0.96, 0.9}
\definecolor{airforceblue}{rgb}{0.36, 0.54, 0.66}
\definecolor{amaranth}{rgb}{0.9, 0.17, 0.31}
\definecolor{babypink}{rgb}{0.96, 0.76, 0.76}
\definecolor{cadetblue}{rgb}{0.37, 0.62, 0.63}
\definecolor{cambridgeblue}{rgb}{0.64, 0.76, 0.68}
\definecolor{carolinablue}{rgb}{0.6, 0.73, 0.89}
\definecolor{darkcyan}{rgb}{0.0, 0.55, 0.55}
\definecolor{etonblue}{rgb}{0.59, 0.78, 0.64}
\definecolor{indianyellow}{rgb}{0.89, 0.66, 0.34}
\definecolor{lilac}{rgb}{0.78, 0.64, 0.78}
\definecolor{pastelred}{rgb}{1.0, 0.41, 0.38}
\definecolor{sandstorm}{rgb}{0.93, 0.84, 0.25}
\definecolor{sunglow}{rgb}{1.0, 0.8, 0.2}

\newcommand{\snippetBox}[2]{\draw[fill=#1,line width=0.5pt] (#2ex,0) rectangle +(1.25ex,1.25ex);}
\useunder{\uline}{\ul}{}

\begin{document}
 
\title{Technical Report: Temporal Aggregate Representations}

\author{Fadime Sener\textsuperscript{1}, Dibyadip Chatterjee\textsuperscript{2}, Angela Yao\textsuperscript{2}\\
\textsuperscript{1}University of Bonn, Germany \\
\textsuperscript{2}National University of Singapore \\
{\tt\small sener@cs.uni-bonn.de, dibyadip@comp.nus.edu.sg, ayao@comp.nus.edu.sg} 
}

\maketitle 

\begin{abstract}

This technical report extends our work presented in ~\cite{sener2020temporal} with more experiments.
In ~\cite{sener2020temporal}, we tackle long-term video understanding, which requires reasoning from current and past or future observations and raises several fundamental questions.
How should temporal or sequential relationships be modelled?
What temporal extent of information and context needs to be processed? 
At what temporal scale should they be derived?
~\cite{sener2020temporal} addresses these questions with a flexible multi-granular temporal aggregation framework.
In this report, we conduct further experiments with this framework on different tasks and a new dataset, EPIC-KITCHENS-100. 
Our code and models can be found in {\url{https://github.com/dibschat/tempAgg}} 

\end{abstract}

\section{Introduction}
\begin{figure*} 
\centering 
 \includegraphics[width=0.80\textwidth]{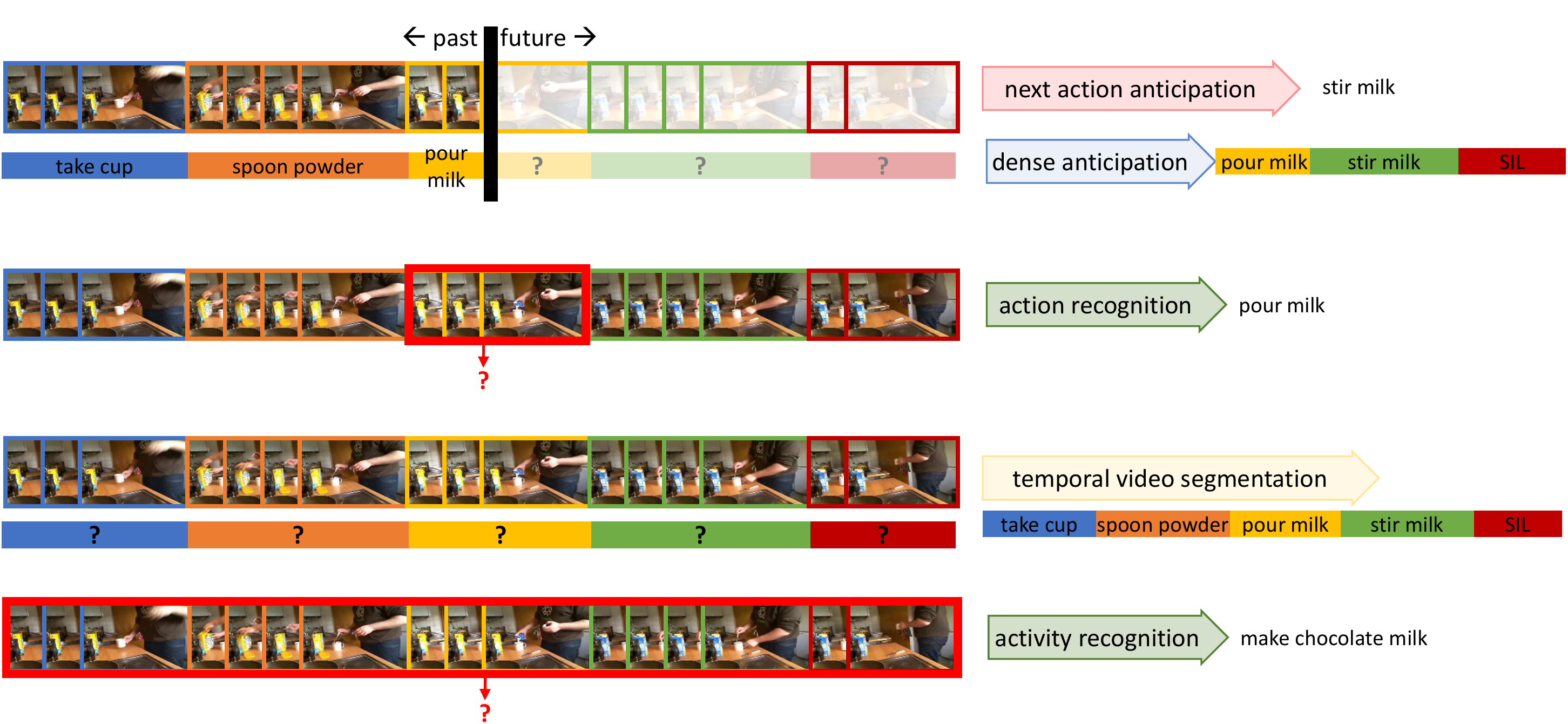}
\caption{
Our temporal aggregates model~\cite{sener2020temporal} is very flexible in that it can be utilized for different tasks easily. 
}
\label{fig:overview_new1}
\end{figure*}

In this work, we utilize the temporal aggregates model presented in~\cite{sener2020temporal} for next action anticipation, action, and activity recognition in long-range videos, see Fig.~\ref{fig:overview_new1}. 
We also test our method on the new EPIC-KITCHENS-100 dataset. 
Our model is described in detail in \cite{sener2020temporal}, and we refer the reader to this paper for further detail.

\begin{figure} 
\centering 
 \includegraphics[width=1.0\columnwidth]{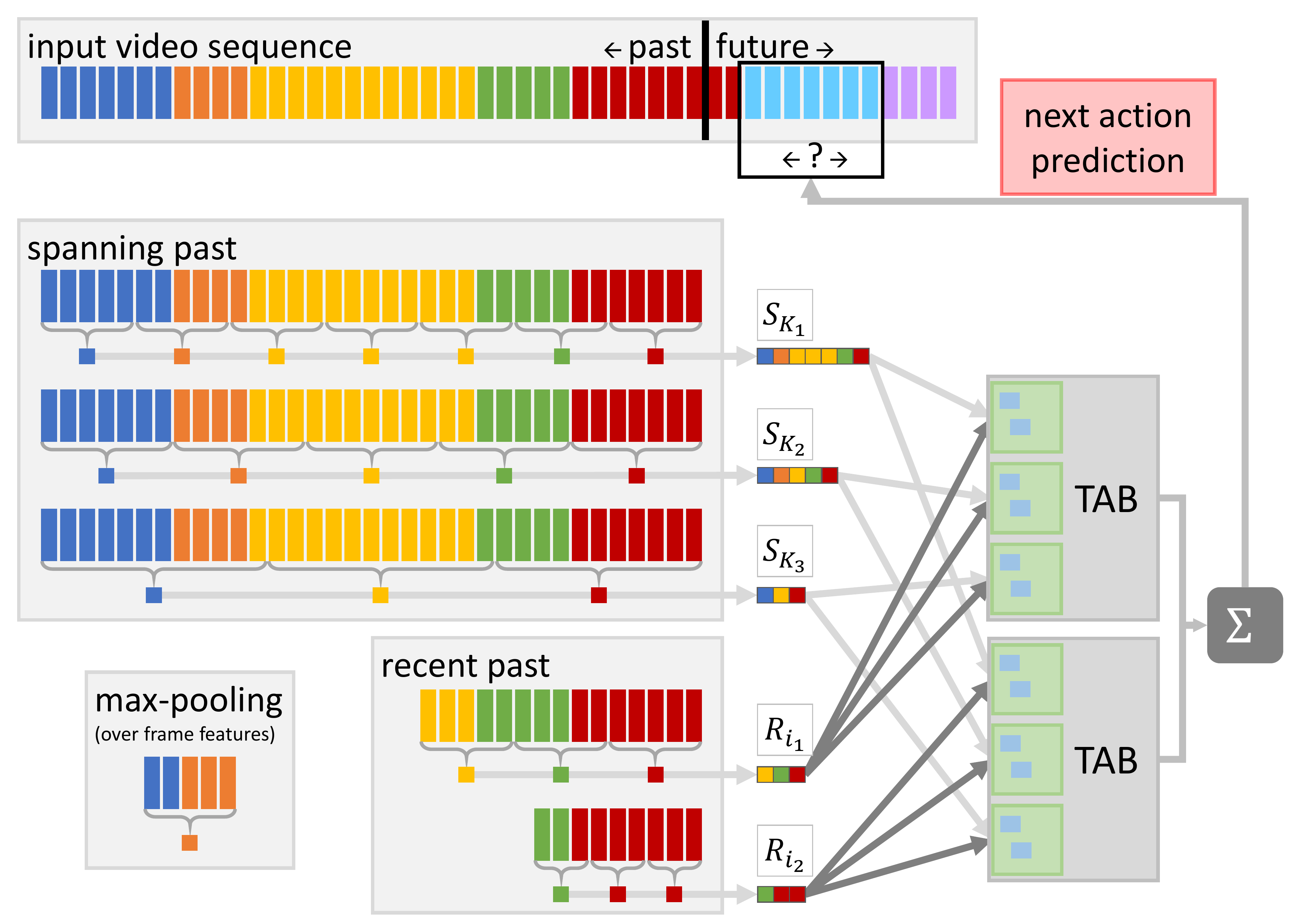}
\caption{
Model overview: In this example we use 3 scales for computing the ``spanning past'' snippet features $\mathbf{S}_{K_1}, \mathbf{S}_{K_2}, \mathbf{S}_{K_3}$, and 2 starting points to compute the ``recent past'' snippet features, $\mathbf{R}_{i_1}, \mathbf{R}_{i_2}$, by max-pooling over the frame features in each snippet. 
Each recent snippet is coupled with all the spanning snippets in our Temporal Aggregation Block (TAB). 
An ensemble of TAB outputs is used for next action anticipation. 
Best viewed in color.
}
\label{fig:overview_new}
\end{figure}

An overview of the building blocks of our temporal aggregates framework can be found in Fig.~\ref{fig:overview_new}. 
We split video streams into snippets of equal length and max-pool the frame features within the snippets. 
We then create ensembles of multi-scale feature representations that are aggregated bottom-up based on scaling and temporal extent. 
Based on different start and end frames $i$ and $j$ and number of snippets $K$, we define two types of snippet features: \emph{`recent'} features $\{\mathcal{R}\}$ from recent observations and \emph{``spanning''} features $\{\mathcal{S}\}$ drawn from the long-term video. 
The recent snippets cover a couple of seconds (or up to a minute, depending on the temporal granularity) before the current time point, while spanning snippets refer to the long-term past and may last up to ten minutes. 
In Fig.~\ref{fig:overview_new} we use two starting points to compute the ``recent past'' snippet features and represent each with $K_R\!=\!3$ number of snippets
(\tikz{\snippetBox{boxYellow}{0}\snippetBox{boxGreen}{1.25}\snippetBox{boxRed}{2.5}} \&
\tikz{\snippetBox{boxGreen}{0}\snippetBox{boxRed}{1.25}\snippetBox{boxRed}{2.5}}).
And we use three scales to compute the ``spanning past'' snippet features with $K = \{7,5,3\}$ (
\tikz{\snippetBox{boxBlue}{0}\snippetBox{boxOrange}{1.25}\snippetBox{boxYellow}{2.5}\snippetBox{boxYellow}{3.75}\snippetBox{boxYellow}{5.0}\snippetBox{boxGreen}{6.25}\snippetBox{boxRed}{7.5}},
\tikz{\snippetBox{boxBlue}{0}\snippetBox{boxOrange}{1.25}\snippetBox{boxYellow}{2.5}\snippetBox{boxGreen}{3.75}\snippetBox{boxRed}{5.0}} \&
\tikz{\snippetBox{boxBlue}{0}\snippetBox{boxYellow}{1.25}\snippetBox{boxRed}{2.5}}
).
Key to both types of representations is the ensemble of snippet features from multiple scales.

Our framework is build in a bottom up manner, starting with the recent and spanning features $\mathcal{R}$ and $\mathcal{S}$, which are coupled with non-local blocks (NLB) within coupling blocks (CB).
Non-local operations~\cite{wang2018non} are applied to capture relationships amongst the spanning snippets and between spanning and recent snippets. 
Two such NLBs are combined in a Coupling Block (CB) which calculates attention-reweighted recent and spanning context representations.
Each recent with all spanning representations are coupled via individual CBs and their outputs are combined in a Temporal Aggregation Block (TAB). 
Outputs of different TABs are then chained together for the task of interest.

\section{Experiments}
\subsection{Implementation Details}
We train our models using the Adam optimizer~\cite{kingma2014adam} with batch size 10, learning rate $10^{-4}$ and dropout rate 0.3.
We train for {\it k} epochs (where {\it k=15} if {\it task=anticipation} \& {\it k=25} if {\it task=recognition}) and decrease the learning rate by a factor of 10 every $10^{\text{th}}$ epoch.
We use 512-D vectors for all non-classification linear layers.

\subsection{Recognizing Long-range Complex Activities}

To validate our model further on a new task, we experiment on classifying long-range complex activities.
Since these videos include multiple actions and are several minutes long, it becomes more challenging to model their temporal structure compared to short-term single action videos, see Fig.~\ref{fig:overview_new1} ``activity recognition''.
Recently, \cite{hussein2019timeception} proposed a neural layer, ``Timeception'', which uses multi-scale temporal-only convolutions for modelling minutes-long complex activity videos, such as ``cooking a meal''. 
Placed on top of backbone CNNs, the permutation invariant convolution layer, PIC \cite{hussein2020pic}, also aims at modelling only the temporal dimension.
PIC is invariant to temporal permutations as it models their correlations regardless of their order, which helps to handle different action orderings in videos.
It also uses pairs of key-value kernels to learn the most representative visual signals in long and noisy videos.

\begin{table}[t]
\centering
\resizebox{\columnwidth}{!}{
\setlength{\tabcolsep}{11pt}
\begin{tabular}{@{}cccc@{}}
\toprule
Dataset & \# spanning scope (s) & $K_R$ & $\{K\}$\\
 \midrule
Breakfast & entire video & 5 & $\{10,15,20\}$\\ 
\bottomrule
\end{tabular}}
\caption{Model parameters for activity recognition on Breakfast.}
\label{tab:my_label_breakfast}
\end{table}

\begin{table}[t]
\centering
\resizebox{\columnwidth}{!}{ 
\setlength{\tabcolsep}{11pt}
\begin{tabular}{|l|l|l|}
\hline
Method & Fine-tuning & Accuracy (\%) \\ \hline
\rowcolor{apricot} I3D & no & 64.3 \\ \hline
\rowcolor{apricot} I3D + Timeception~\cite{hussein2019timeception} & no & 69.3 \\ \hline
\rowcolor{apricot} I3D + ours & no & \textbf{80.8} \\ \hline \hline
\rowcolor{olivine!45} I3D & yes & 80.6 \\ \hline
\rowcolor{olivine!45}I3D + Timeception~\cite{hussein2019timeception} & yes & 86.9 \\ \hline
\rowcolor{olivine!45}I3D+ PIC~\cite{hussein2020pic} & yes & \textbf{89.8} \\ \hline
\end{tabular}}
\vspace{1pt}
\caption{
Comparisons to methods developed for recognizing long-range complex activities, Timeception~\cite{hussein2019timeception} and PIC~\cite{hussein2020pic} on the Breakfast Actions dataset.
Our method outperforms Timeception~\cite{hussein2019timeception} by a significant margin showing the superiority of our method in modelling long-range activities.
}
\label{chapter5:tab:longrangeComp}
\end{table}

We experiment on the Breakfast actions dataset~\cite{kuehne2014language}, which contains 1712 videos of 10 complex activities such as ``making coffee''. 
In our model, we divide videos into three partitions and use each partition as a recent snippet. 
We use the entire video for computing our spanning snippets. 
The model parameters are presented in Table~\ref{tab:my_label_breakfast}.

We report our comparisons in Table~\ref{chapter5:tab:longrangeComp} on Breakfast actions using two types of I3D features, where one is the features from an I3D model trained on Kinetics only, and the other is the features from an I3D model fine-tuned on the Breakfast dataset.
Our method outperforms Timeception~\cite{hussein2019timeception} by 11.4\%, and the I3D backbone by 16.5\%.
\cite{hussein2020pic} use the fine-tuned I3D features on Breakfast and shows a 3.1\% improvement over Timeception~\cite{hussein2019timeception}.
Fine-tuning improves the accuracy by 16.3\% and shows that there is room for improvement for our method using better feature representations.

\begin{table*}[!h]
\centering
\resizebox{\textwidth}{!}{
\setlength{\tabcolsep}{11.5pt}
\begin{tabular}{@{}ccccccc@{}}
\toprule
Task & \# segments & $\{i\}/\{i, j\}$(in seconds (s)) & spanning scope (s) & $K_R$ & $\{K\}$\\
\midrule
Anticipation & 90K & $\{t\!-\!1.6,t\!-\!1.2,t\!-\!0.8,t\!-\!0.4\}$ & 6 & 2 & $\{2,3,5\}$\\
Recognition & 90K & $\{(s, e),(s - 1, e + 1), (s - 2, e + 2), (s - 3, e + 3)\}$ & ($s-6$, $e+6$) & 5 & $\{2,3,5\}$\\
\bottomrule
\end{tabular}}
\caption{
Dataset details and our respective model parameters for anticipation and recognition.
s and e refers to the start and end times of the segments for action recognition.
}
\label{tab:my_label}
\end{table*}

\begin{table*}[!t]
\centering
\resizebox{\textwidth}{!}{{}
\setlength{\tabcolsep}{11.1pt}
\begin{tabular}{@{}cclccccccccccc@{}}
\toprule
& & & \multicolumn{3}{c}{Overall} & & \multicolumn{3}{c}{Unseen Participants} & & \multicolumn{3}{c}{Tail Classes} \\ \cmidrule(lr){4-6} \cmidrule(lr){8-10} \cmidrule(l){12-14}
Split & Modality & & Verb & Noun & Act. & & Verb & Noun & Act. & & Verb & Noun & Act. \\ \midrule
\multirow{5}{*}{\rotatebox[origin=c]{90}{\bf Val}} 
& RGB & & {\ul \textbf{24.22}} & {\ul 29.76} & {\ul 13.02} & & 27.04 & 22.95 & 12.21 & & {\ul \textbf{16.23}} & {\ul \textbf{22.93}} & {\ul 10.41} \\
& Flow & & 18.90 & 18.68 & 7.27 & & 26.53 & 18.86 & 9.54 & & 10.65 & 12.53 & 5.25 \\
& Obj & & 20.45 & 27.64 & 10.45 & & 24.17 & {\ul 24.71} & 11.45 & & 12.55 & 19.31 & 7.36 \\
& ROI & & 21.22 & 26.61 & 11.62 & & 25.49 & 19.16 & 10.10 & & 13.36 & 19.91 & 9.10 \\ \cmidrule(l){2-14} 
& Fusion & & 23.15 & \textbf{31.37} & \textbf{14.73} & & \textbf{28.01} & \textbf{26.23} & \textbf{14.47} & & 14.50 & 22.47 & \textbf{11.75} \\ \midrule
\rotatebox[origin=c]{90}{\bf Test} & Fusion & & \textbf{21.76} & \textbf{30.59} & \textbf{12.55} & & \textbf{17.86} & \textbf{27.04} & \textbf{10.46} & & \textbf{13.59} & \textbf{20.62} & \textbf{8.85} \\
\bottomrule
\end{tabular}}
\caption{
Action {\bf anticipation} results (class-mean top-5 recall) on EPIC-KITCHENS-100 validation and test sets.
We report our results for RGB, Flow, Obj and ROI modalities and the late fusion of the predictions from all these modalities (Fusion).
}
\label{tab:epic_ex_ant_test}
\end{table*}

\begin{table*}[!t]
\centering
\resizebox{\textwidth}{!}{{}
\setlength{\tabcolsep}{6.1pt}
\begin{tabular}{@{}cccccccccccccccccc@{}}
\toprule
& & & \multicolumn{7}{c}{Overall} & & \multicolumn{3}{c}{Unseen Participants} & & \multicolumn{3}{c}{Tail Classes} \\ \cmidrule(lr){4-10} \cmidrule(lr){12-14} \cmidrule(l){16-18} 
& & & \multicolumn{3}{c}{Top-1 Accuracy (\%)} & & \multicolumn{3}{c}{Top-5 Accuracy (\%)} & & \multicolumn{3}{c}{Top-1 Accuracy (\%)} & & \multicolumn{3}{c}{Top-1 Accuracy (\%)} \\ \cmidrule(lr){4-6} \cmidrule(lr){8-10} \cmidrule(lr){12-14} \cmidrule(l){16-18} 
Split & Modality & & Verb & Noun & Act. & & Verb & Noun & Act. & & Verb & Noun & Act. & & Verb & Noun & Act. \\ \midrule
\multirow{5}{*}{\rotatebox[origin=c]{90}{\bf Val}} & RGB & & 59.92 & 45.14 & 36.87 & & 86.72 & 71.09 & 56.97 & & 47.51 & 31.46 & 27.51 & & 26.82 & {\ul 21.89} & {\ul 18.04} \\
& Flow & & {\ul 62.81} & 37.51 & 32.84 & & 87.70 & 63.22 & 53.17 & & {\ul 53.80} & 32.86 & {\ul 28.26} & & {\ul 27.23} & 8.00 & 12.72 \\
& Obj & & 49.89 & 41.70 & 30.97 & & 84.02 & {\ul 73.56} & 54.00 & & 48.36 & {\ul 34.84} & 27.63 & & 23.64 & 20.63 & 14.27 \\
& ROI & & 57.75 & 43.16 & 35.48 & & 86.36 & 70.87 & 56.65 & & 47.32 & 34.65 & 26.76 & & 26.42 & 20.16 & 16.94 \\ \cmidrule(l){2-18} 
& Fusion & & \textbf{66.00} & \textbf{53.35} & \textbf{45.26} & & \textbf{89.39} & \textbf{80.40} & \textbf{66.93} & & \textbf{56.62} & \textbf{44.60} & \textbf{38.12} & & \textbf{30.57} & \textbf{25.26} & \textbf{22.42} \\ \midrule
\rotatebox[origin=c]{90}{\bf Test} & Fusion & & \textbf{62.68} & \textbf{51.66} & \textbf{42.65} & & \textbf{86.28} & \textbf{73.85} & \textbf{64.05} & & \textbf{56.25} & \textbf{46.33} & \textbf{36.06} & & \textbf{24.99} & \textbf{18.00} & \textbf{15.92} \\ \bottomrule
\end{tabular}}
\caption{
Action {\bf recognition} results on EPIC-KITCHENS-100 validation and test sets.
We report our results for modalities RGB, Flow, Obj and ROI and late fusion of the predictions from all these modalities (Fusion).
}
\label{tab:epic_ex_rec_test}
\end{table*}

\subsection{Experiments on EPIC-KITCHENS-100}
\textbf{Epic-Kitchens}-100~\cite{damen2020rescaling} is the recently released extension to Epic-Kitchens-55~\cite{damen2018scaling}.
It is the largest egocentric dataset with 100 hours of egocentric recordings capturing participants' daily kitchen activities with a head-mounted camera. 
There are around $90$K pre-trimmed segments extracted from 700 long videos. Each segment is annotated with an action composed of a verb and noun classes, \eg, \emph{``pour water''}. 
There are 4,025 actions composed of 97 verbs and 300 nouns. 
The dataset provides RGB and optical flow images, as well as bounding boxes extracted by a hand-object detection framework~\cite{shan2020understanding}.

\paragraph{Parameters:} The spanning scales $\{K\}$, recent scale $K_R$, recent starting points $\{i\}$ and recent ending points $\{j\}$ are given in Table~\ref{tab:my_label}.
In our work, we anticipate or recognize the action classes directly rather than anticipating or recognizing the verbs and nouns independently~\cite{damen2018scaling} which is shown to outperform the latter~\cite{furnari2018leveraging}.
We use the training and validation sets provided by~\cite{damen2020rescaling} for selecting our model parameters.

\subsubsection{Features}
We use the appearance (RGB), motion (optical flow), and object-based features provided by ~\cite{furnari2019rulstm} for reporting the baseline results on EPIC-100.
They independently train two CNNs using the TSN ~\cite{wang2016temporal} framework on RGB and flow images for action recognition on EPIC-Kitchens-100. 
~\cite{furnari2019rulstm} also trains object detectors to recognize the 352 object classes of the EPIC-KITCHENS-100 dataset.

We additionally extract regions of interest (ROI) features from this pre-trained TSN model (on RGB), provided by ~\cite{furnari2019rulstm}, for the hand-object interaction regions in frames. 
We use the interacting hand-object bounding boxes provided by~\cite{shan2020understanding} and consider the union of these boxes to be our ROI for each frame.
The RGB features from this ROI help our model ignore the background clutter, which adversely affects our performance as it focuses primarily on interacting regions.
The feature dimensions are 1024, 1024, and 352, 1024 for appearance, motion, object, and ROI features, respectively.

\subsubsection{Anticipation on EPIC-KITCHENS-100}\label{sec:daily_ant}
The anticipation task of EPIC-KITCHENS-100 requires anticipating the future action $\tau_{\alpha}\!=\!1$s before it starts. 
We train our model separately for each feature modality (appearance, motion, object and RoI) with the parameters described in Table~\ref{tab:my_label} and apply late fusion to the predictions from all these modalities by average voting to compute our final results.

Table \ref{tab:epic_ex_ant_test} summarizes our results (class-mean top-5 recall (\%)) for validation and hold-out test sets on EPIC-KITCHENS-100 for all (overall) and unseen participants and tail classes. 
Overall the ensemble of all modalities improves action scores for overall, unseen and tail classes while training our model solely on RGB performs better for verb and noun scores. 
Our submission in the challenge leaderboard~\footnote{Test results obtained by submission to {\url{https://competitions.codalab.org/competitions/25925\#results}}} is named as ``temporalAgg''.

\subsubsection{Recognition on EPIC-KITCHENS-100}\label{sec:daily_rec} 
For recognition, we classify pre-trimmed action segments.
We train our model separately for each feature modality using the model parameters described in Table~\ref{tab:my_label}. 
During inference, similar to our anticipation, a late fusion of the predictions from modalities RGB, Flow, Obj, and ROI is used.

Following the EPIC-KITCHENS-100 protocol~\cite{damen2020rescaling}, we report Top-1/5 accuracies on both the validation and test sets for all (overall) and unseen participants and tail classes in Table \ref{tab:epic_ex_rec_test}. 
Fusing all modalities improves all scores significantly for all evaluation categories. 
Our submission in the challenge leaderboard~\footnote{Test results obtained by submission to {\url{https://competitions.codalab.org/competitions/25923\#results}}} is ``temporalAgg''.

{\small
\bibliographystyle{ieee_fullname}
\bibliography{egbib}
}

\end{document}